\documentclass{article}

\usepackage{microtype}
\usepackage{graphicx}
\usepackage{booktabs} %

\usepackage{miller}
\usepackage{ulem}

\usepackage{hyperref}

\usepackage[accepted]{synsml2023}

\usepackage{amsmath}
\usepackage{amssymb}
\usepackage{mathtools}
\usepackage{cuted}
\usepackage{flushend}
\usepackage{amsthm}

\usepackage[capitalize,noabbrev]{cleveref}

\theoremstyle{plain}

\theoremstyle{definition}

\theoremstyle{remark}

\usepackage[textsize=tiny]{todonotes}

\everypar{\looseness=-1}
\synsmltitlerunning{Simulation-based Inference with the Generalized Kullback-Leibler Divergence}

\begin{document}

\twocolumn[
\synsmltitle{Simulation-based Inference with the Generalized Kullback-Leibler Divergence}

\synsmlsetsymbol{equal}{*}

\begin{synsmlauthorlist}
\synsmlauthor{Benjamin Kurt Miller}{ams}
\synsmlauthor{Marco Federici}{ams}
\synsmlauthor{Christoph Weniger}{ams}
\synsmlauthor{Patrick Forr\'e}{ams}
\end{synsmlauthorlist}

\synsmlaffiliation{ams}{University of Amsterdam}

\synsmlcorrespondingauthor{Benjamin Kurt Miller}{b.k.miller@uva.nl}

\synsmlkeywords{Machine Learning Simulation-based Inference Kullback-Leibler Divergence}

\vskip 0.3in
]

\printAffiliationsAndNotice{}  %

\begin{abstract}
In Simulation-based Inference, the goal is to solve the inverse problem when the likelihood is only known implicitly. Neural Posterior Estimation commonly fits a normalized density estimator as a surrogate model for the posterior. This formulation cannot easily fit unnormalized surrogates because it optimizes the Kullback-Leibler divergence. We propose to optimize a generalized Kullback-Leibler divergence that accounts for the normalization constant in unnormalized distributions.
The objective recovers Neural Posterior Estimation when the model class is normalized and unifies it with Neural Ratio Estimation, combining both into a single objective. We investigate a hybrid model that offers the best of both worlds by learning a normalized base distribution and a learned ratio. We also present benchmark results.
\end{abstract}

\section{Simulation-based Inference}
\label{introduction}

Consider this motivating example: Your task is to infer the mass ratio of a binary black hole system $\bthetao$ from observed gravitational wave strain data $\bxo$ of their merger. Numerical simulation can map from hypothetical mass ratio $\btheta$ to simulated gravitational wave strain data $\bx$ using general relativity, but the inverse map is unspecified and intractable. \emph{Simulation-based Inference} (\SBI) approaches this problem probabilistically \citep{Cranmer2020, sisson2018handbook}.

Although we cannot evaluate the density, we assume the simulator samples from conditional distribution $p(\bx \mid \btheta)$. Once we specify a prior $p(\btheta)$, the inverse amounts to estimating posterior $p(\btheta \mid \bxo)$ where $\btheta$ represents simulator input parameters and $\bx$ the simulated output observation. In our \emph{amortized} approach, we learn a surrogate model $q(\btheta \mid \bx)$ that approximates the posterior for any $\bx \sim p(\bx)$, which we assume includes $\bxo$, while limiting excessive simulation.

\begin{figure}[t]
    \vskip 0.2in
    \begin{center}
    \centerline{\includegraphics[width=\columnwidth]{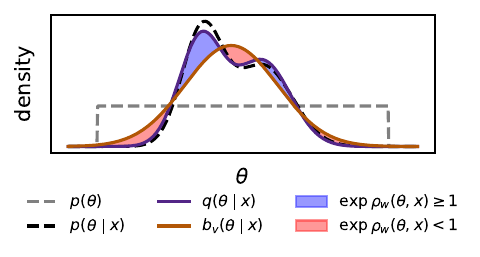}}
    \caption{
        A proposed posterior surrogate model $q(\btheta \mid \bx)$ consisting of two components: (1) An easy-to-sample-from base distribution $b_{v}(\btheta \mid \bx)$ that approximates the posterior $p(\btheta \mid \bx)$ better than the prior $p(\btheta)$. (2) A density ratio estimated by a flexible energy-based model $\exp \rho_{w}$ that reduces the density of the base distribution when $\exp \rho_{w}(\btheta, \bx) < 1$ and increases it when $\exp \rho_{w}(\btheta, \bx) \geq 1$. We train the surrogate by minimizing the generalized Kullback-Leibler divergence. Rejection sampling is easy as $b_{v}$ is close to $q$.
    }
    \label{conceptual-unnormalized-distribution}
    \end{center}
    \vskip -0.2in
\end{figure}

\subsection{Limitations of Simulation-based Inference Methods}

Neural Posterior Estimation (\NPE) \citep{papamakarios2016fast} learns a surrogate posterior by solving an optimization problem for normalized $\tilde q(\btheta \mid \bx)$ in model class $\tilde \Fcal$:
\begin{align}
    \label{npe-objective}
    \begin{aligned}
    \tilde q^*(\btheta \mid \bx) &\in \argmin_{\tilde q(\btheta \mid \bx) \in \Fcal}
    \E_{p(\bx)}\lB \KL(p(\btheta \mid \bx) \Mid \tilde q(\btheta \mid \bx)) \rB  \\
    &= \argmin_{\tilde q(\btheta \mid \bx) \in \Fcal} \E_{p(\bx,\btheta)}\lB - \ln \tilde q(\btheta \mid \bx) \rB,
    \end{aligned}
\end{align}
where $\KL$ denotes the Kullback-Leibler divergence. Although practical \citep{dax2021real}, likelihood-based \NPE suffers from model choice limitations. The conditional distribution is restricted to inflexible distributions parameterized by Mixture Density Networks \citep{bishop1994mixture} or Normalizing Flows \citep{papamakarios2019normalizing} that require special consideration for multi-modality \cite{huang2018neural, cornish2020relaxing} and high-dimensionality \cite{kong2020expressive, reyesgonzalez2023testing}.

Methods that perform \NPE, but with an unnormalized surrogate $q(\btheta \mid \bx)$ have recently been developed. \citet{ramesh2021gatsbi} proposed an adversarial objective in their method GATSBI, but training can be unstable \citep{salimans2016improved}. 
There has also been work on score-based training using sequential proposals \citep{sharrock2022sequential} or a flexible number of observations $\bx_o^{(1)}, \ldots \bx_o^{(N)}$ \citep{geffner2022score}. These methods require Langevin dynamics for sampling and it is non-trivial to evaluate their (unnormalized) density.

Neural Ratio Estimation (\NRE) \citep{thomas2016likelihood, Hermans2019, durkan2019neural, miller2022contrastive} approximates the likelihood-to-evidence ratio $\frac{p(\bx \mid \btheta)}{p(\bx)}$. It can fit marginals \cite{miller2021truncated}, proving useful in practice \cite{cole2021fast, bhardwaj2023peregrine}. However, it is not a variational estimate of the posterior \cite{poole2019variational} and suffers from saturation effects \cite{rhodes2020telescoping}.

\paragraph{Alternatives}
We focus on \NPE and \NRE, but other approximation techniques exist. Approximate Bayesian Computation employs a similarity kernel between summary statistics of simulations $T(\bx)$ and an observation $T(\bx_o)$ to draw samples from an approximate posterior \citep{sisson2018handbook}.

Estimating the likelihood $p(\bx \mid \btheta)$ is another approach \citep{wood2010statistical, papamakarios2019sequential, pacchiardi2022score}, but it requires modeling the complex generative process and sampling may be non-trivial. \citet{glaser2022maximum} propose maximum likelihood estimation to learn an unnormalized, energy-based model for $p(\bx \mid \btheta)$. It is similar to our proposed Posterior-to-Prior Ratio surrogate; however, \citet{glaser2022maximum} use a particle approximation of the gradient of the log partition function for training while we avoid this step by minimizing our proposed objective \eqref{our-objective} that does not contain the log partition function.

\subsection{One Objective, Three Models: General Surrogates}

We overcome the aforementioned issues with \NPE by proposing to optimize the Generalized KL-Divergence instead. It enables fitting (unnormalized) surrogate models based on either: a normalized density model, a posterior-to-prior ratio, or a hybrid model with advantages of both. 
The first case recovers \NPE and the hybrid model is visualized in \cref{conceptual-unnormalized-distribution}. Hybrid models have generative applications in energy-based modeling \citep{arbel2021generalized} and can help reduce the variance in estimating mutual information \citep{federici2023effectiveness}.

We derive the Generalized KL-Divergence from the so-called $\varphi$-divergences \citep{renyi1961measures, csiszar1963eine, ali1966general}. Our objective, the divergence between conditionals taken in expectation over $p(\bx)$, is identified with a lower bound to the average KL-Divergence called \emph{
Tractable Unnormalized version of the Barber and Agakov lower bound on Mutual Information} \cite{poole2019variational, barber2003algorithm, nguyen2010estimating, nowozin2016f, belghazi2018mine}.

\paragraph{Contribution}
We provide a unification of two methods, \NPE and \NRE, by proposing the average Generalized KL-divergence as an objective for \SBI. This formulation enables training a hybrid model, which is novel to \SBI. We support the objective and hybrid model with benchmark results.

\section{Generalized Kullback-Leibler Divergence}
\label{sec:generalized-kld}

\begin{Def}[The Generalized KL-divergence]
Let $p(\btheta)$ and $q(\btheta)$ be (unnormalized) probability distributions. We define the \emph{Generalized KL-divergence of $q(\btheta)$ w.r.t.\ $p(\btheta)$} by:
\begin{align}
   &\GKL(p(\btheta) \Mid q(\btheta )) \coloneqq \int \phi\lp \frac{q(\btheta)}{p(\btheta)}\rp p(\btheta)\,d\btheta, \\
   &\phi\lp r \rp \coloneqq - \ln r + r - 1.
\end{align}
\end{Def}

The central properties of the Generalized KL-divergence are that (a) Gibbs' inequality: $\GKL(p(\btheta) \Mid q(\btheta)) \ge 0$ holds, even in the case of unnormalized probability distributions, and (b) $\GKL(p(\btheta) \Mid q(\btheta)) = 0$ if and only if $p(\btheta) = q(\btheta)$ $p(\btheta)$-almost surely. These properties imply that we can optimize the objective over a flexible model class (unnormalized distributions) using the variational principle. The divergence is general because for normalized $p(\btheta)$, $q(\btheta)$ it reduces to the original KL-divergence. Proof in \cref{gkld-details}. 

Let $Z_p \coloneqq \int p(\btheta) \, d\btheta$ and 
$Z_q \coloneqq \int q(\btheta) \, d\btheta$
be normalizing constants. When $\GKL(p(\btheta)\|q(\btheta)) = 0$, $p(\btheta) = q(\btheta)$ and $Z_p = Z_q$. When $\KL(p(\btheta)/Z_p \| q(\btheta)/Z_q)=0$, $p(\btheta)/Z_p = q(\btheta)/Z_q$, but we do \emph{not necessarily} have $Z_p = Z_q$! (Both are $p(\btheta)$-almost surely.) In this way $\GKL$, is ``stronger''. We present an inequality between these divergences:
\begin{align}
    \label{renormalized-kl-inequality}
    Z_p \cdot \KL(p(\btheta)/Z_p \Mid q(\btheta)/Z_q) \le \GKL(p(\btheta) \Mid q(\btheta)).
\end{align}
Proof in \cref{gkld-details}.

\begin{figure}[bt]
    \vskip 0.2in
    \begin{center}
    \centerline{\includegraphics[width=\columnwidth]{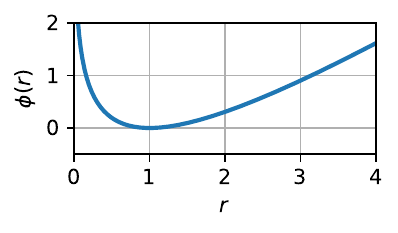}}
    \caption{
        The integrand $\phi$ in the Generalized Kullback-Leibler Divergence. Its non-negativity leads to Gibbs' equality in the objective on (unnormalized) probability distributions and the optimum is unique because $\phi(r)=0$ only when $r=1$, i.e., when $p = q$.
    }
    \label{generalized-kullback-leibler-curve}
    \end{center}
    \vskip -0.2in
\end{figure}

\subsection{Application to Simulation-based Inference}
Assume we have a fixed simulator given by the conditional distribution $p(\bx \mid \btheta)$; 
we can sample from the joint distribution $\btheta_i \sim p(\btheta)$, $\bx_i \sim p(\bx \mid \btheta_i)$; and we aim to learn the posterior distribution $p(\btheta \mid \bx)$ for any $\bx$ within the support of $p(\bx)$. We thus aim to solve this minimization problem, using some model class $\Fcal$ of (unnormalized) distributions:
\begin{align}
    \label{optimization-problem}
    q(\btheta \mid \bx) \in \argmin_{q(\btheta \mid \bx) \in \Fcal} \E_{p(\bx)} \lB \GKL\lp p(\btheta \mid \bx) \| q(\btheta \mid \bx) \rp \rB
\end{align}
The objective function, given the above modelling choices, can be simplified according to \cref{our-objective} where $C$ denotes all terms not dependent on $q(\btheta \mid \bx)$.
\begin{figure*}[bt]
    \begin{align}
    \begin{aligned}
        \label{our-objective}
       \E_{p(\bx)} \lB \GKL \lp p(\btheta \mid \bx) \Mid q(\btheta \mid \bx)\rp \rB
       &= \iint \lp 
        -\ln q(\btheta \mid \bx) + \ln p(\btheta \mid \bx) + \frac{q(\btheta \mid \bx)}{p(\btheta \mid \bx)} -1 \rp \, p(\btheta,\bx)\,d\btheta \,d\bx \\
       &= \E_{p(\btheta, \bx)}\lB - \ln q(\btheta \mid \bx)  \rB 
       + \E_{p(\bx)}\lB \int q(\btheta \mid \bx) \, d\btheta  \rB + C
    \end{aligned}
    \end{align}
\end{figure*}
The objective is estimated and optimized on mini-batches of samples, but the exact formula depends on how we choose to model $q(\btheta \mid \bx)$. 

We present three options for $q(\btheta \mid \bx)$: A normalized density estimator, an energy-based model that estimates the posterior-to-prior ratio $\frac{p(\btheta \mid \bx)}{p(\btheta)}$, and a hybrid model that uses a normalized density as a base distribution and an energy-based model to fit a posterior-to-base-distribution ratio. The first of these techniques is exactly \NPE from \SBI; the second produces a ratio, similar to \NRE, but uses the Generalized KL-divergence objective; and the hybrid is novel within \SBI.

\paragraph{Normalized Density Surrogate}

Consider the parameterization $q(\btheta \mid \bx) \coloneqq b_{v}(\btheta \mid \bx)$ where $b_{v}$ is a normalized density estimator like a normalizing flow, or normal distribution with weights $v$. In this case, $\E_{p(\bx)} \lB \int q(\btheta \mid \bx) \, d\btheta \rB = 1$, so it becomes part of $C$ and is no longer involved in optimization. Here the objective becomes identical to \NPE, like \cref{npe-objective}. We sample $m = 1, \ldots, M$ data points as follows:
\begin{align*}
    \btheta_m &\sim p(\btheta), & \bx_m &\sim p(\bx \mid \btheta_m),
\end{align*}
to estimate the loss function as
\begin{align}
    \label{loss-normalized-density}
    \frac{1}{M} \sum_{m=1}^{M} \lB - \ln b_{v}(\btheta_m \mid \bx_m) \rB.
\end{align}
Drawing samples $\hat\btheta \sim q(\btheta \mid \bxo)$ from this model is as simple as sampling from the density estimator.

\paragraph{Posterior-to-Prior Energy-based Ratio Surrogate}

Consider the parameterization $q(\btheta \mid \bx) \coloneqq \exp(\rho_{w}(\btheta,\bx)) \cdot p(\btheta)$ where $\rho_{w}$ is a scalar, parametric function, like a neural network, with weights $w$. Since this surrogate is not necessarily normalized, we must consider all terms in \cref{our-objective} for optimization, except $C$. We sample additional data like so:
\begin{align*}
    \btheta_m' &\sim p(\btheta), & \bx_m' &\sim p(\bx \mid \btheta_m'),
\end{align*}
and combine with above samples to approximate the loss
\begin{align}
    \label{loss-ratio}
    \frac{1}{M}\sum_{m=1}^M \lB 
    - \rho_w(\btheta_m,\bx_m)
    + \exp\lp \rho_w(\btheta_m,\bx_m') \rp  \rB.
\end{align}
$\btheta_m'$ is merely used to sample $\bx_m'$; it does not appear in the objective directly. $\bx_m'$ may bootstrapped by permuting index $m$. Accurately estimating $\E_{p(\bx)} \lB Z_{w}(\bx) \rB \coloneqq \E_{p(\bx)} \lB  \int \exp(\rho_{w}(\btheta, \bx)) p(\btheta) \, d\btheta \rB$ may require many samples from $p(\btheta) p(\bx)$; however, we used a single sample $(\btheta_m', \bx_m')$ as an unbiased estimate. This inaccuracy may contribute to the low-quality fit observed in Section~\ref{experiments}; however, investigation is left for future work.

Since estimates of the log-partition function are biased in maximum likelihood training of energy-based models \cite{glaser2022maximum}, the gradient of the log-partition function $\nabla_{w} \E_{p(\bx)} \lB \log Z_{w}(\bx) \rB$ is approximated by sampling \citep[Equation (4)]{song2021train}. Since the log-partition function does not appear in our objective \eqref{our-objective}, we can do an unbiased Monte Carlo estimate of $\E_{p(\bx)} \lB  Z_{w}(\bx) \rB$ and take gradients using automatic differentiation. Without an improved proposal, as in the hybrid surrogate, this term in our objective has high variance \cite{federici2023effectiveness}.

This surrogate is generally not normalized and drawing samples $\hat\btheta \sim q(\btheta \mid \bxo)$ requires additional computation. In low dimensions, rejection sampling from $p(\btheta)$ can be tractable; otherwise, Markov-chain Monte Carlo (\MCMC) becomes necessary. We did \MCMC to draw samples in \cref{experiments}.

\paragraph{Hybrid Surrogate}

Consider parameterizing $q(\btheta \mid \bx) \coloneqq \exp(\rho_{w}(\btheta,\bx)) \cdot b_{v}(\btheta \mid \bx)$. This surrogate is not necessarily normalized. We propose to estimate the relevant term in \cref{our-objective} using Monte Carlo samples from the normalized base distribution. We simplify the term suggestively:
\begin{align*}
    \E_{p(\bx)} \lB \int q(\btheta \mid \bx) \, d\btheta \rB = \E_{b_{v}(\btheta \mid \bx) \, p(\bx)} \lB \exp(\rho_{w}(\btheta, \bx)) \rB.
\end{align*}
We found that taking gradients on both $\rho_{w}$ and $b_{v}$ did not facilitate learning. Instead, take samples $\tilde\btheta_m \sim b_{v}(\btheta \mid \bx_m)$ \emph{without} applying the reparameterization trick: Given parametric invertible function $f_{b_{v}}$ of normalizing flow $b_{v}$ then,
\begin{align*}
    \epsilon_m &\sim \Ncal(\epsilon \mid 0, I), &
    \tilde\btheta_m &:= f_{b_{v}}(\epsilon; \bx_m), & \nabla_{v} f_{b_{v}} &\coloneqq 0.
\end{align*}
This amounts to first fitting the base distribution  for one gradient step $b_{v}$, followed by fitting the log ratio $\rho_{w}$. Given the data points sampled above, we estimate the loss function
\begin{align}
    \label{loss-hybrid}
    \begin{aligned}
    \frac{1}{M} \sum_{m=1}^M \Bigg[ -\ln b_{v}(\btheta_m \mid \bx_m) &- \rho_{w}(\btheta_m, \bx_m) \\ 
    &+ \exp(\rho_{w}(\tilde\btheta_m, \bx_m)) \Bigg].
    \end{aligned}
\end{align}

We estimate $\E_{p(\bx)} \lB  \int \exp(\rho_{w}(\btheta, \bx)) b_{v}(\btheta \mid \bx) \, d\btheta \rB$ using a single sample $(\tilde\btheta_m, \bx_m)$, similarly to the \emph{ratio} surrogate.

When sampling $\hat\btheta \sim q(\btheta \mid \bxo)$, we leverage the $b_{v}(\btheta \mid \bx)$ distribution as a proposal and perform rejection sampling according to $\exp(\rho_{w}(\btheta, \bx))$. Since $b_{v}(\btheta \mid \bx)$ is close to $q(\btheta \mid \bx)$, this results in a tractable percentage of accepted samples. It was effective for all experiments in \cref{experiments}, but high-dimensional surrogates may require alternatives.

\section{Experiments}
\label{experiments}

\begin{figure*}[tb]
    \vskip 0.2in
    \begin{center}
    \centerline{\includegraphics[width=\textwidth]{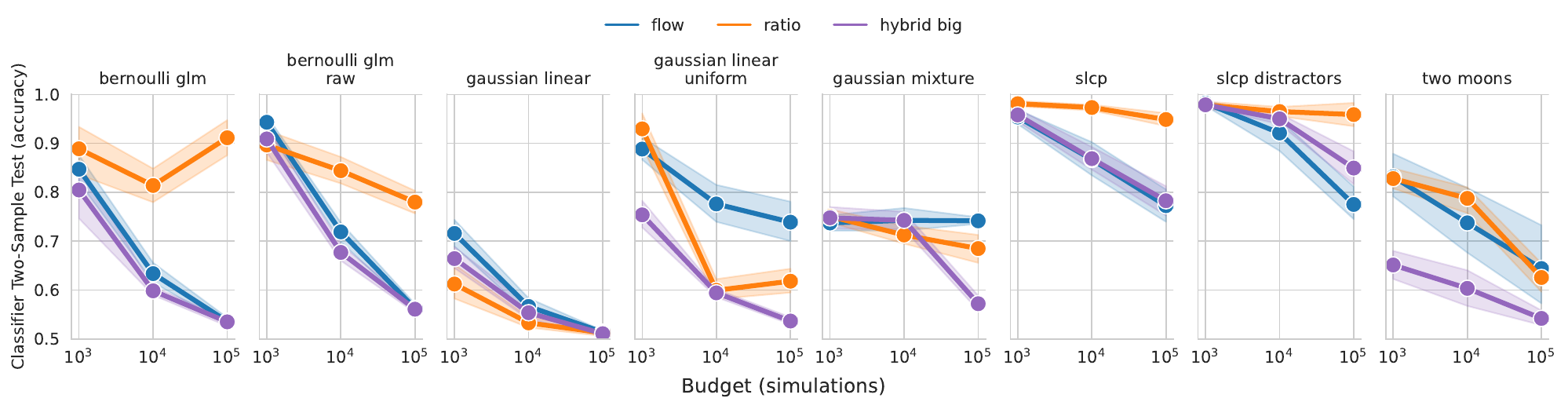}}
    \caption{
        The C2ST accuracy and 95\% confidence intervals are plotted versus simulation budget for three surrogate models on eight benchmark tasks. We estimate the exactness of the surrogate model against tractable posterior samples. Lower values indicate either a better fit, or limitations of the C2ST itself. The \emph{flow} architecture trains a Normalized Density Surrogate Model that optimizes \cref{loss-normalized-density} and is equivalent to \NPE. The \emph{ratio} architecture trains the Posterior-to-Prior Energy-based Ratio Surrogate Model that optimizes \cref{loss-ratio}. Finally, the \emph{hybrid big} architecture trains a Hybrid Surrogate Model that optimizes \cref{loss-hybrid}.
    }
    \label{c2st_vs_samples}
    \end{center}
    \vskip -0.2in
\end{figure*}

It has become standard in the \SBI literature to measure the exactness of the surrogate against a tractable posterior as a function of simulation budget, despite failing to represent the practitioner's setting \citep{hermans2021averting}. \citet{sbibm} collected ten priors and simulators, each with ten parameter-observation pairs and 10,000 samples from the corresponding likelihood-based posterior, to create the so-called Simulation-based Inference Benchmark. The parameters range between two- and ten-dimensional while the simulations range between two- and 100-dimensional. The benchmark measures the five-fold cross-validated Classifier Two-Sample Test (C2ST) \cite{friedman2003multivariate, lopez2017revisiting} accuracy by comparing samples from the posterior and the surrogate at simulation budgets between $10^{3}$ and $10^{5}$ joint samples. Classification accuracies of 0.5 indicate that either the surrogate is indistinguishable from the posterior from the given samples, or that the classifier does not have the capacity to tell the distributions apart. p-values and E-values \cite{pandeva2022valuating} are not considered. 

Experiments were done with a Neural Spline Flow (NSF)-based normalized density surrogate \cite{durkan2019neural}, a posterior-to-prior ratio-based surrogate, and a hybrid surrogate which trained a ratio against a Masked Autoregressive Flow (MAF)-based density estimator \cite{papamakarios2017masked}. Following \cite{delaunoy2023balancing}, we appended an unconditional bijection from the final distribution layer to the prior support in all normalizing flows. We found that it increased training stability when the prior $p(\btheta)$ was uniform.

\paragraph{Results}
We report C2ST results across 8 tasks, all aforementioned models, and several architectures in \cref{c2st_vs_samples}. We averaged over ten random seeds per plotted point to create the 95\% confidence intervals with an approximately 76 node-day computational cost for all runs across all hyperparameters. Architecture and training details are in \cref{experimental-details}, including \cref{c2st_vs_samples_comprehensive} that shows additional results for \emph{ratio big} and \emph{hybrid} that correspond to the same model, but with varied neural network hyperparameter choices. Additionally, for the \emph{hybrid} models, we report the C2ST for samples from either the base distribution $b_{v}(\btheta \mid \bx)$ or the full hybrid model $\exp \rho_{w}(\btheta, \bx) b_{v}(\btheta \mid \bx)$ in \cref{c2st_vs_samples_hybrid}. It diagnoses how much each component contributes to the overall surrogate. 

The \emph{hybrid big} model was generally more accurate than \emph{flow} or \emph{ratio}, although there were exceptions: In Gaussian Linear and SLCP Distractors the \emph{ratio} or \emph{flow} models were better. \emph{Ratio} was very sensitive to neural network size and cannot be trusted to solve Gaussian Linear accurately with arbitrary neural network design. SLCP has a complex shape and may have benefited from using a Neural Spline Flow as the base distribution for \emph{hybrid big}. Two Moons features extreme multi-modality allowing \emph{hybrid big} to shine: The base distribution $b_{v}(\btheta \mid \bx)$ left a typical narrow tail connecting each moon, but the ratio $\exp \rho_{w}(\btheta, \bx)$ erased this tail density. A visualization of an example surrogate and reference posterior can be found in Figure~\ref{two_moons_posteriors}.

\section{Conclusion}

We proposed to use the Generalized KL-divergence, in expectation over $p(\bx)$, as an objective for \SBI, connected it to \NPE, proposed estimating the posterior-to-prior ratio using this objective, and proposed a natural hybrid model class for \SBI. We evaluated the exactness of fits empirically on eight benchmark problems at three simulation budgets, representing over two months of computation time. Our conclusion was that the increased flexibility of our objective and the \emph{hybrid} model was generally beneficial in comparison to \emph{flow}, i.e. \NPE, but especially for Two Moons that features a multi-modal posterior. Fitting the \emph{ratio} alone, with our objective, was very sensitive to neural network hyperparameter choices, emphasizing the importance of the hybrid model.

In the hybrid model, one must choose a variational family for the base distribution $b_{v}(\btheta \mid \bx)$. The base distribution must be more tightly concentrated than the prior to see improvement in performance, while also covering the entire posterior mass. In scientific settings, a conditional normal distribution with mean and covariance estimated by neural networks should be effective in most situations (without long-tailed posteriors). For the benchmark, we choose MAF since it was flexible and unlikely to exclude posterior mass.

In Section~\ref{sec:generalized-kld} and our experiments, we used a single sample to estimate $\E_{p(\bx)} \lB Z_{w}(\bx) \rB$. As a Monte Carlo estimate, it has a variance $\propto \frac{1}{N}$ where $N$ represents the number of samples in the estimate. The constant of proportionality may be very large, meaning that a single sample is insufficiently accurate. Investing the quantitative effect of the number of contrastive samples on learning has been left to future work, although we expect that it might have a significant effect for the \emph{ratio} surrogate. A theoretical analysis of the variance of $\E_{p(\bx)} \lB Z_{w}(\bx) \rB$ is offered in \citet{federici2023effectiveness} along with experiments in estimating the mutual information using a varied number of so-called ``negative'' samples.

A natural follow-up work would extend our method to the so-called \emph{sequential} case, where we train the surrogate estimator in a sequence of rounds. In each round, simulation data is drawn such that the current posterior estimate focuses attention on $\bx$ which are ``close'' to an observation-of-interest $\bxo$. We plan to use the flexibility of our objective by updating the sampling distribution $p_n(\btheta, \bx)$ across $n$ rounds.

\section*{Broader impact}

The primary application of \SBI is to solve the inverse problem on observations using high-fidelity simulation data. The broader societal impact is therefore limited to which simulators are considered for application. 

Since \SBI does not rely on likelihoods, it can be challenging to determine whether surrogates are overfit and provide inaccurate certainty about estimated parameters.
We emphasize rigorous statistical testing to confirm results from \SBI to avoid drawing inaccurate conclusions.

\section*{Acknowledgements}

Benjamin Kurt Miller is part of the ELLIS PhD program.
Christoph Weniger received funding from the European Research Council (ERC) under the European Union’s Horizon 2020 research and innovation programme (Grant agreement No. 864035 -- UnDark).

\nocite{lueckmann2017flexible, greenberg2019automatic, glockler2021variational, ramesh2021gatsbi, tran2017hierarchical, talts2018validating, zhao2021diagnostics, masserano2022simulation, linhart2022validation, alsing2018massive, alsing2019fast, gratton2017glass, lemos2023sampling, dalmasso2021likelihood, liese2006divergences}

\bibliography{bibliography}
\bibliographystyle{synsml2023}

\newpage
\appendix
\onecolumn

\section{Generalized Kullback-Leibler Divergence Details}
\label{gkld-details}

We continue directly from \cref{sec:generalized-kld} with a formulation of Gibbs' inequality as a Theorem. Since the Generalized KL-Divergence is a $\phi$-divergence, this is review for our specific divergence choice.

\begin{Thm}[Gibbs' inequality for the Generalized KL-Divergence]
We always have the inequality:
\begin{align}
    \GKL(p(\btheta) \Mid q(\btheta)) & \ge 0.
\end{align}
Furthermore, the equality $\GKL(p(\btheta) \Mid q(\btheta)) =0$ holds if and only if $p(\btheta)=q(\btheta)$ for $p(\btheta)$-almost-all points $\btheta$, in other words, if they are equal inside the support of $p(\btheta)$.
\end{Thm}
\begin{proof}
   Consider the following function $\phi:\, (0,\infty) \to \R$ given by:
\begin{align}
    \label{eqn:sckld-integrand}
    \phi(r) &:= -\ln(r) + r -1,
\end{align}
with the additional setting $\phi(0):=\phi(\infty):=\infty$. 
It always holds that $\phi(r) \ge 0$, with equality if and only if $r=1$.
So, for any non-negative measurable function $R \ge 0$ we have: $\int \phi(R(\bz))\, p(\bz) \, d\bz \ge 0$ with equality if and only if $R(\bz)=1$ $p(\bz)$-almost-surely. 
Our case follows from $\bz=\btheta$, $R(\bz)=\frac{q(\btheta)}{p(\btheta)}$ and $p(\bz)=p(\btheta)$.
\end{proof}

\begin{Rem}
\begin{enumerate}
    \item The above equality holds for unnormalized probability distributions, not just up to normalizing constants.
    \item For normalized probability distributions $p_1(\btheta)$ and $p_2(\btheta)$ we recover the classical KL-divergence:
    \begin{align}
        \GKL(p_1(\btheta) \Mid p_2(\btheta)) &= \E_{p_1(\btheta)}\lB \ln \frac{p_1(\btheta)}{p_2(\btheta)} \rB =: \KL(p_1(\btheta) \Mid p_2(\btheta)).
    \end{align}
     In this sense, the Generalized $\KL$-divergence is a real generalization of the classical $\KL$-divergence.
\end{enumerate}
\end{Rem}

Our general Gibbs' inequality now allows us to formulate a generalization of the classical \emph{minimal (relative) entropy principle} \citep{kullback1951information, shore1980axiomatic, cover1999elements}:

\begin{Pri}[The principle of minimal Generalized KL-divergence]
Let $p(\btheta)$ be an underlying ``true,'' given probability distribution that we want to approximate with a(n unnormalized) probability distribution $q(\btheta)$ from a certain model class $\Fcal$, expressing certain prior knowledge or constraints. 

Then the \emph{principle of minimal Generalized KL-divergence} expresses that one should choose that $q(\btheta)$ from $\Fcal$ that has minimal Generalized KL-divergence to $p(\btheta)$, i.e.\ one should choose:
\begin{align}
    q^*(\btheta) \in \argmin_{q(\btheta) \in \Fcal}
    \GKL(p(\btheta) \Mid q(\btheta)).
\end{align}
\end{Pri}

Finally, we prove \cref{renormalized-kl-inequality}: 
\begin{align}
\begin{aligned}
    Z_p \cdot \KL(p(\btheta)/Z_p \Mid q(\btheta)/Z_q) &=
    Z_p \int \frac{p(\btheta)}{Z_p} \ln \frac{p(\btheta) / Z_p}{q(\btheta) / Z_q} \, d\btheta \\
    &= -\int p(\btheta) \ln \frac{p(\btheta)}{q(\btheta)} \, d\btheta + Z_p \ln \frac{Z_q}{Z_p}  \\
    &\leq -\int p(\btheta) \ln \frac{p(\btheta)}{q(\btheta)} \, d\btheta + Z_p \lp \frac{Z_q}{Z_p} - 1 \rp \\
    &= -\int p(\btheta) \ln \frac{p(\btheta)}{q(\btheta)} \, d\btheta + Z_q - Z_p \\ 
    &= -\int p(\btheta) \ln \frac{p(\btheta)}{q(\btheta)} \, d\btheta + \int q(\btheta) \, d\btheta - \int p(\btheta) \, d\btheta \\
    &= -\int p(\btheta) \lp \ln \frac{p(\btheta)}{q(\btheta)} + \frac{q(\btheta)}{p(\btheta)} - 1 \rp \, d\btheta \\
    &= \GKL(p(\btheta) \Mid q(\btheta)).
\end{aligned}
\end{align}

\section{Experimental Details}
\label{experimental-details}

In this section we include more information about the tasks, hyperparameters that we chose, and a few more results.

\subsection{Simulation-based Inference Benchmark Task Details}
We provide a short summary of all of the inference tasks we consdiered from the \SBI benchmark by \citet{sbibm}.

\begin{itemize}[align=left]
	\item[\textbf{Bernoulli GLM}]  This task is a generalized linear model. The likelihood is Bernoulli distributed. The data is a 10-dimensional sufficient statistic from an 100-dimensional vector. The posterior is 10-dimensional with only one mode.
	\item[\textbf{Bernoulli GLM Raw}]  This is the same task as above, but instead the entire 100-dimensional observation is shown to the inference method rather than the summary statistic.
	\item[\textbf{Gaussian Linear}]  A simple task with a Gaussian distributed prior and a Gaussian likelihood over the mean. Both have a $\Sigma=0.1 \cdot I$ covariance matrix. The posterior is also Gaussian. It is performed in 10-dimensions for the observations and parameters.
	\item[\textbf{Gaussian Linear Uniform}]  This is the same as the task above, but instead the prior over the mean is a 10-dimensional uniform distribution from -1 to 1 in every dimension.
	\item[\textbf{Gaussian Mixture}]  This task occurs in the ABC literature often. Infer the common mean of a mixture of Gaussians where one has covariance matrix $\Sigma=1.0 \cdot I$ and the other $\Sigma=0.01 \cdot I$. It occurs in two dimensions.
	\item[\textbf{SLCP}]  A task which has a very simple non-spherical Gaussian likelihood, but a complex posterior over the five parameters which, via a non-linear function, define the mean and covariance of the likelihood. There are five parameters each with a uniform prior from -3 to 3. The data is four-dimensional but we take two samples from it.
	\item[\textbf{SLCP with Distractors}] This is the same as above but instead the data is concatenated with 92 dimensions of Gaussian noise.
	\item[\textbf{Two Moons}]  This task exhibits a crescent shape posterior with bi-modality--two of the attributes often used to stump \MCMC samplers. Both the data and parameters are two dimensional. The prior is uniform from -1 to 1.
\end{itemize}

\subsection{Hyperparameters}

In this section we report the hyperparameters for each of our models in \cref{hyperparameters}. AdamW is an optimizer introduced by \citet{loshchilov2017decoupled}. LWCR stands for Linear Warmup Cosine Annealing. NSF stands for Neural Spline Flow and MAF stands for Masked Autoregressive Flow.

\begin{table}[htb]
\centering
\caption{Hyperparameters}
\label{hyperparameters}
\begin{tabular}{llllll}
\toprule
Model Name & flow & ratio & ratio big & hybrid & hybrid big \\
\midrule
Batch size & 16384 & 16384 & 16384 & 16384 & 16384 \\
Embedding Net & ResNet &  &  & ResNet & ResNet \\
Embedding Net Hidden Dim & [64] &  &  & [64] & [64] \\
Embedding Net Activation & gelu &  &  & gelu & gelu \\
Embedding Net Normalization & Layer Norm &  &  & Layer Norm & Layer Norm \\
Flow & NSF &  &  & MAF & MAF \\
Flow Num Transforms & 5 &  &  & 5 & 5 \\
Flow Num Bins & 8 &  &  &  &  \\
Flow Hidden Dim & [64] &  &  & [64] & [64] \\
Flow Activation & relu &  &  & relu & relu \\
Ratio Estimator &  & ResNet & ResNet & ResNet & ResNet \\
Ratio Estimator Hidden Dim &  & [128] & [256, 256] & [128] & [256, 256] \\
Ratio Estimator Activation &  & gelu & gelu & gelu & gelu \\
Ratio Estimator Normalization &  & Layer Norm & Layer Norm & Layer Norm & Layer Norm \\
Optimizer & AdamW & AdamW & AdamW & AdamW & AdamW \\
Learning Rate & 0.001 & 0.001 & 0.001 & 0.001 & 0.001 \\
Weight Decay & 0.001 & 0.001 & 0.001 & 0.001 & 0.001 \\
amsgrad & True & True & True & True & True \\
LR Schedule & LWCA & LWCA & LWCA & LWCA & LWCA \\
LR Schedule Warmup Percent & 0.100 & 0.100 & 0.100 & 0.100 & 0.100 \\
LR Schedule Starting LR & 1e-8 & 1e-8 & 1e-8 & 1e-8 & 1e-8 \\
LR Schedule eta & 1e-8 & 1e-8 & 1e-8 & 1e-8 & 1e-8 \\
Early Stopping & True & True & True & True & True \\
Early Stop Minimum Delta & 0.003 & 0.003 & 0.003 & 0.003 & 0.003 \\
Early Stop Patience & 322 & 322 & 322 & 322 & 322 \\
\bottomrule
\end{tabular}
\end{table}

\subsection{Further Results}

\begin{figure}[htb]
    \vskip 0.2in
    \begin{center}
    \centerline{\includegraphics[width=\textwidth]{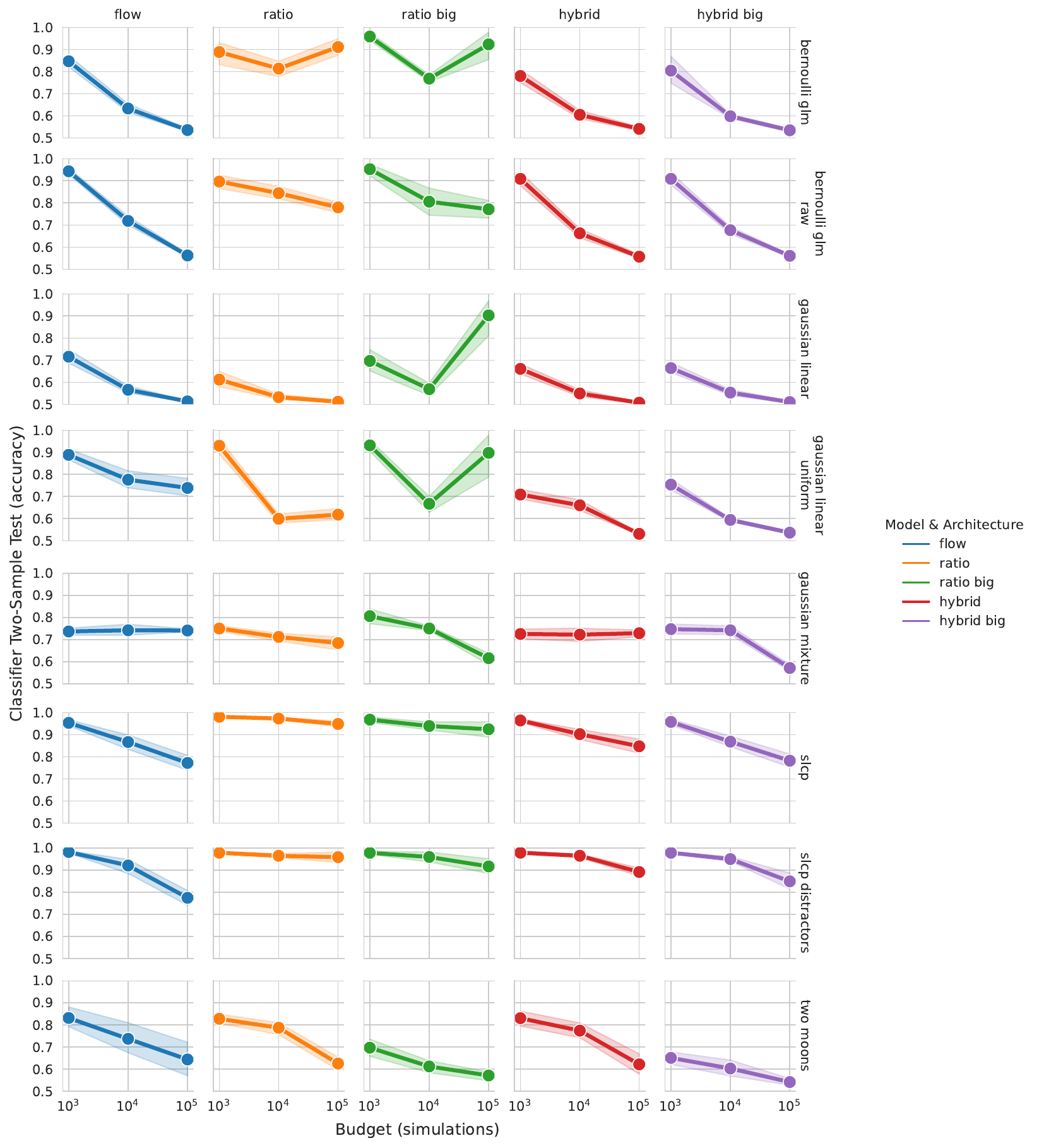}}
    \caption{
        The C2ST accuracy and 95\% confidence intervals are plotted versus simulation budget for five surrogate models on eight benchmark tasks. We estimate the exactness of the surrogate model against tractable posterior samples. Lower values indicate either a better fit, or limitations of the C2ST itself. Details about model hyperparameters can be found in \cref{hyperparameters}.
    }
    \label{c2st_vs_samples_comprehensive}
    \end{center}
    \vskip -0.2in
\end{figure}

\begin{figure}
    \vskip 0.2in
    \centering
    \begin{subfigure}[b]{0.45\textwidth}
     \centering
     \includegraphics[width=\textwidth]{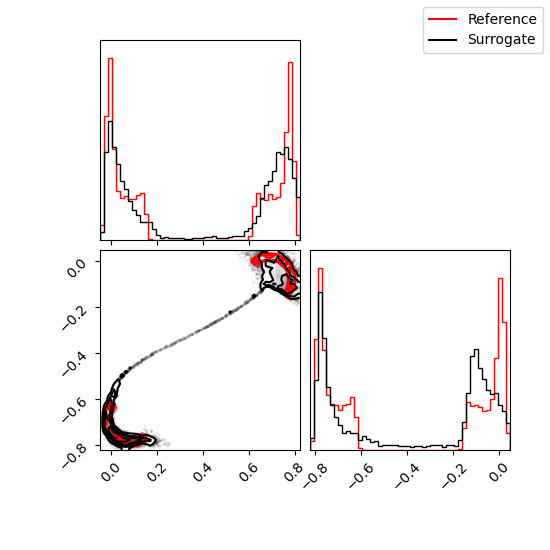}
    \end{subfigure}
    \hfill
    \begin{subfigure}[b]{0.45\textwidth}
     \centering
     \includegraphics[width=\textwidth]{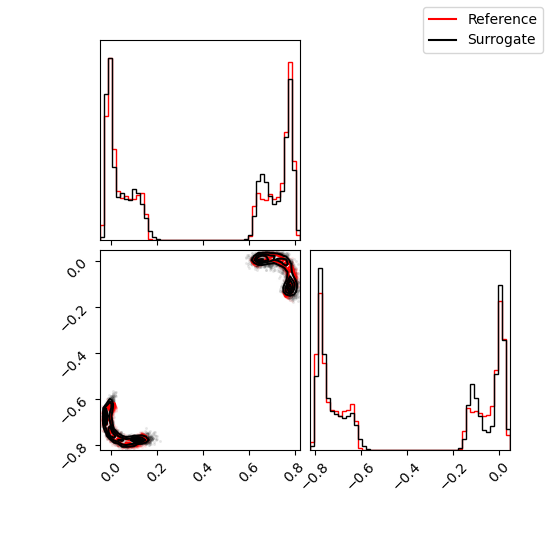}
    \end{subfigure}
    \caption{
        Two corner plots of the posterior and surrogate densities for Two Moons $\bx_{9}$. In both figures, the same reference posterior is drawn in red. The black lines represent different parts of the \emph{same} hybrid surrogate. On the left, the samples come from density estimator $b_{v}(\btheta \mid \bx)$. On the right, samples come from the full hybrid model $\exp \rho_{w}(\btheta, \bx) b_{v}(\btheta \mid \bx)$. The ratio estimator improves the shape significantly.
    }
    \label{two_moons_posteriors}
    \vskip 0.2in
\end{figure}

\begin{figure}[htb]
    \vskip 0.2in
    \begin{center}
    \centerline{\includegraphics[width=\textwidth]{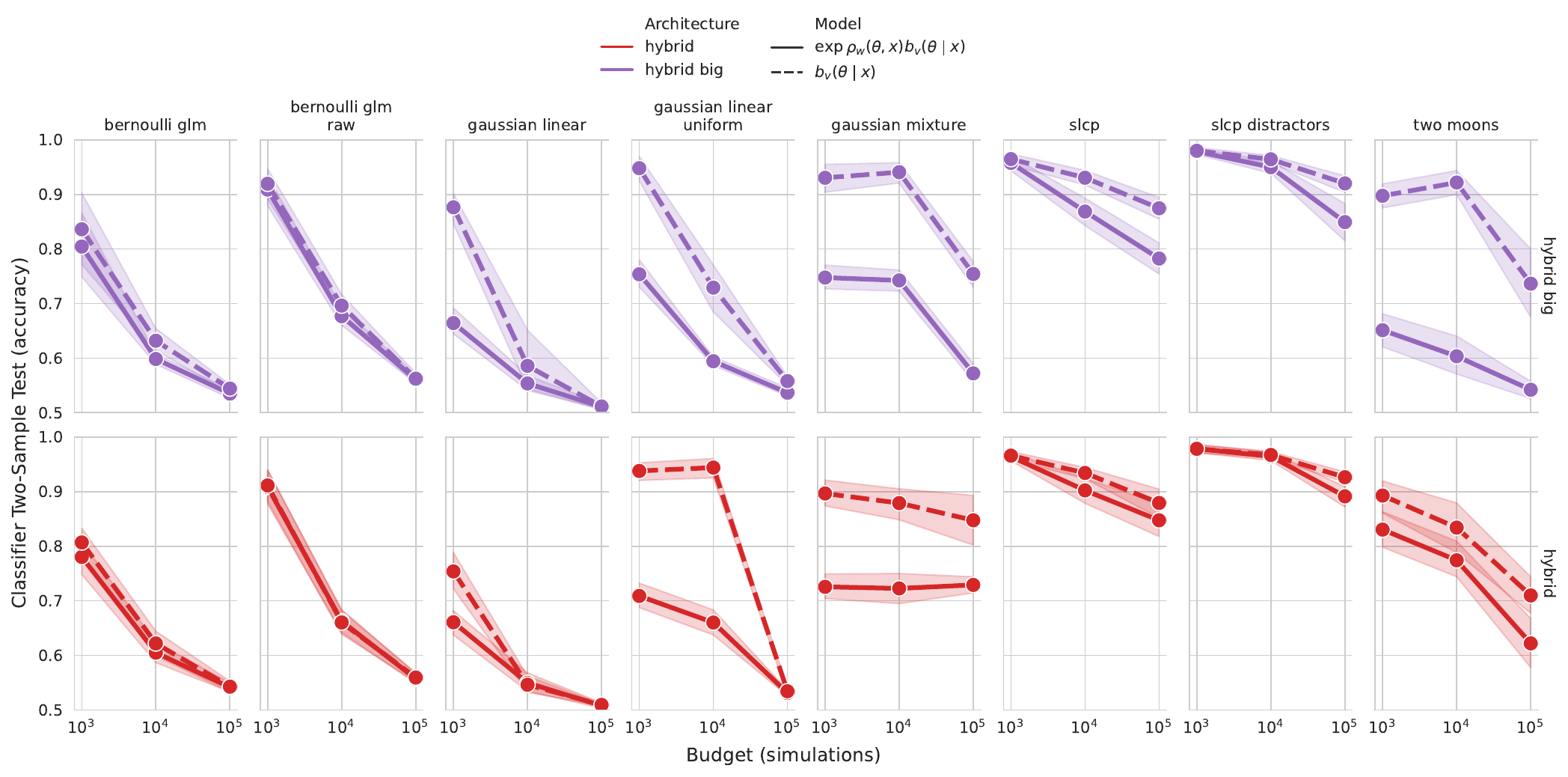}}
    \caption{
        The C2ST accuracy and 95\% confidence intervals are plotted versus simulation budget for two hybrid surrogate models on eight benchmark tasks. We estimate the exactness of the surrogate model against tractable posterior samples. The analysis was performed on samples from the \emph{same} hybrid model, but the dotted line represents samples from only the normalized conditional density estimator $b_{v}(\btheta \mid \bx)$ while the solid line represents samples from the full hybrid surrogate $\exp \rho_{w}(\btheta, \bx) b_{v}(\btheta \mid \bx)$. Lower values indicate either a better fit, or limitations of the C2ST itself. Details about model hyperparameters can be found in \cref{hyperparameters}.
    }
    \label{c2st_vs_samples_hybrid}
    \end{center}
    \vskip -0.2in
\end{figure}

We present results from the \emph{ratio big} and \emph{hybrid} models, along with repeated presentation of previous results, in \cref{c2st_vs_samples_comprehensive}. 
We break the hybrid model into parts and show results based on taking samples directly from the underlying normalized base distribution $b_{v}(\btheta \mid \bx)$ and compare that to samples from the full hybrid model $\exp \rho_{w}(\btheta, \bx) b_{v}(\btheta \mid \bx)$. Qualitative results on a observation $\bx_{9}$ from Two Moons is plotted in \cref{two_moons_posteriors}, and quantitative results across tasks are plotted in \cref{c2st_vs_samples_hybrid}.

\end{document}